\documentclass[letterpaper, 10 pt, conference]{ieeeconf}  

\IEEEoverridecommandlockouts                              

\usepackage{graphics} 
\usepackage{epsfig} 
\usepackage{times} 
\usepackage{amsmath} 
\usepackage{amssymb}  
\usepackage{lipsum}
\usepackage{color}
\usepackage{graphicx}
\usepackage{algorithm,algorithmic}
\usepackage{subcaption}
\usepackage{setspace}
\usepackage{wrapfig}
\usepackage[export]{adjustbox}
\usepackage[bottom]{footmisc}
\usepackage{xr}
\usepackage{hyperref}

\title{\LARGE \bf
Learning Exploration Strategies to Solve Real-World Marble Runs
}
\author{Alisa Allaire$^{1}$ and Christopher G. Atkeson$^{1}$
\thanks{*This material is based upon work supported by the National Science Foundation under Grant IIS-1849287}
\thanks{Robotics Institute, Carnegie Mellon University, Pittsburgh, PA, USA {\tt\small \{aallaire, cga\}@cmu.edu} }%
}

\begin{document}

\maketitle
\thispagestyle{empty}
\pagestyle{empty}

\begin{abstract}

Tasks involving locally unstable or discontinuous dynamics (such as bifurcations and collisions) remain challenging in robotics, because small variations in the environment can have a significant impact on task outcomes. For such tasks, learning a robust deterministic policy is difficult. We focus on structuring exploration with multiple stochastic policies based on a mixture of experts (MoE) policy representation that can be efficiently adapted. The MoE policy is composed of stochastic sub-policies that allow exploration of multiple distinct regions of the action space (or strategies) and a high-level selection policy to guide exploration towards the most promising regions. We develop a robot system to evaluate our approach in a real-world physical problem solving domain. After training the MoE policy in simulation, online learning in the real world demonstrates efficient adaptation within just a few dozen attempts, with a minimal sim2real gap. Our results confirm that representing multiple strategies promotes efficient adaptation in new environments and strategies learned under different dynamics can still provide useful information about where to look for good strategies. 

\end{abstract}

\section{INTRODUCTION}

Developing intelligent systems with the efficient and flexible physical reasoning capabilities of humans remains one of the greatest challenges in robotics. Tasks involving locally unstable or discontinuous dynamics (such as bifurcations and collisions) are particularly difficult because small, possibly unobservable, variations in the environment can have a significant impact on the task outcomes. We are inspired by prior work that proposes simulation-based mechanical puzzles as benchmarks for physical reasoning \cite{allen2020tools, bakhtin2019phyre}. These puzzles often involve locally unstable and discontinuous dynamics, emphasizing collisions as multiple objects move and interact over extended periods of time. Unlike most tasks addressed using reinforcement learning, actions can only be taken at the start of the task (setting the initial configuration of objects). There is no possibility of further control or re-planning after the objects start to move. Success depends on reasoning about the task outcome based on the initial state. 

\begin{figure}[t!]
    \centering
    \begin{subfigure}[t!]{0.68\linewidth}
         \centering
         \includegraphics[width=\linewidth,valign=t]{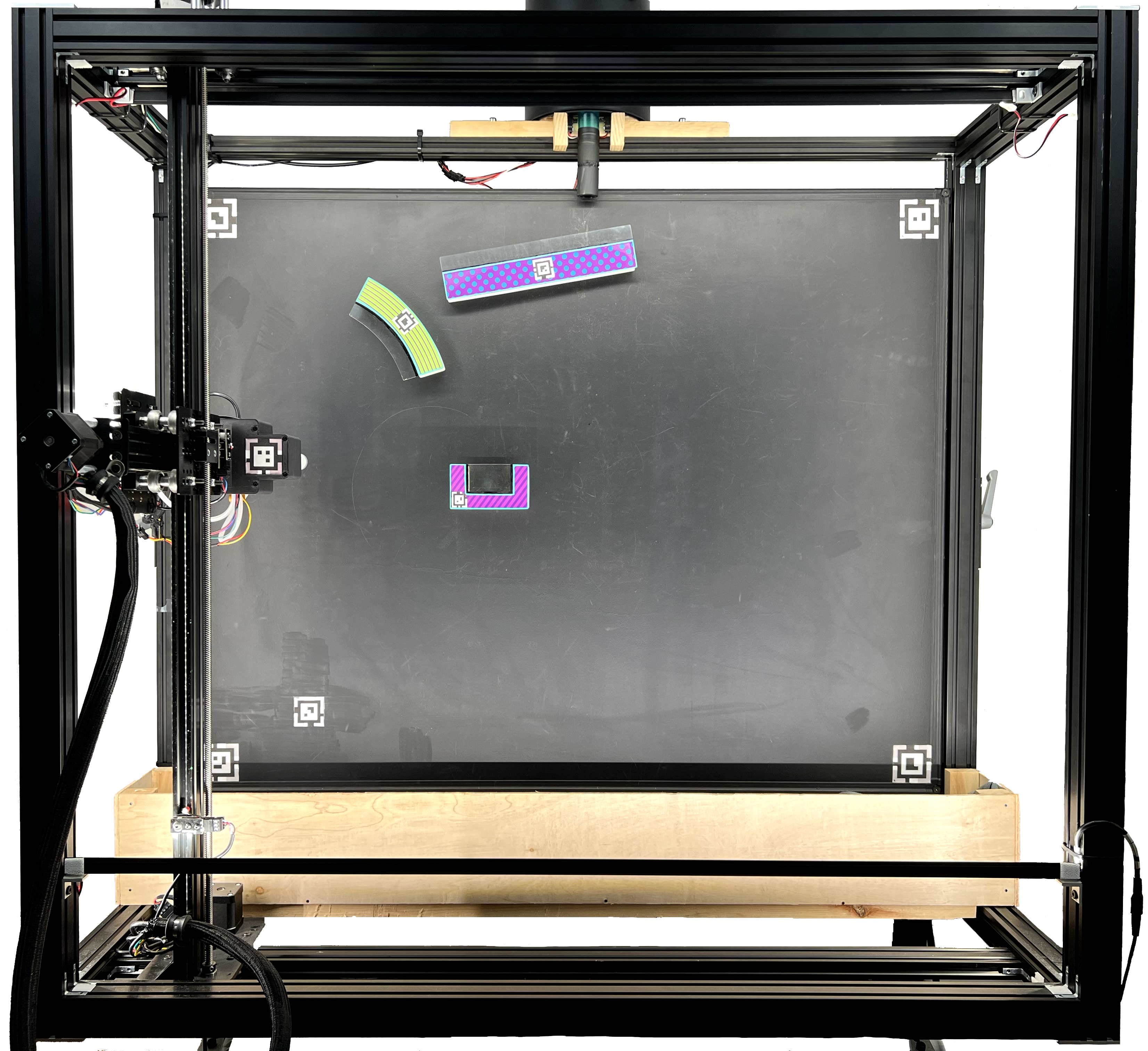}
    \end{subfigure}
    \begin{subfigure}[t!]{0.3\linewidth}
        \centering
        \includegraphics[width=\linewidth,valign=t]{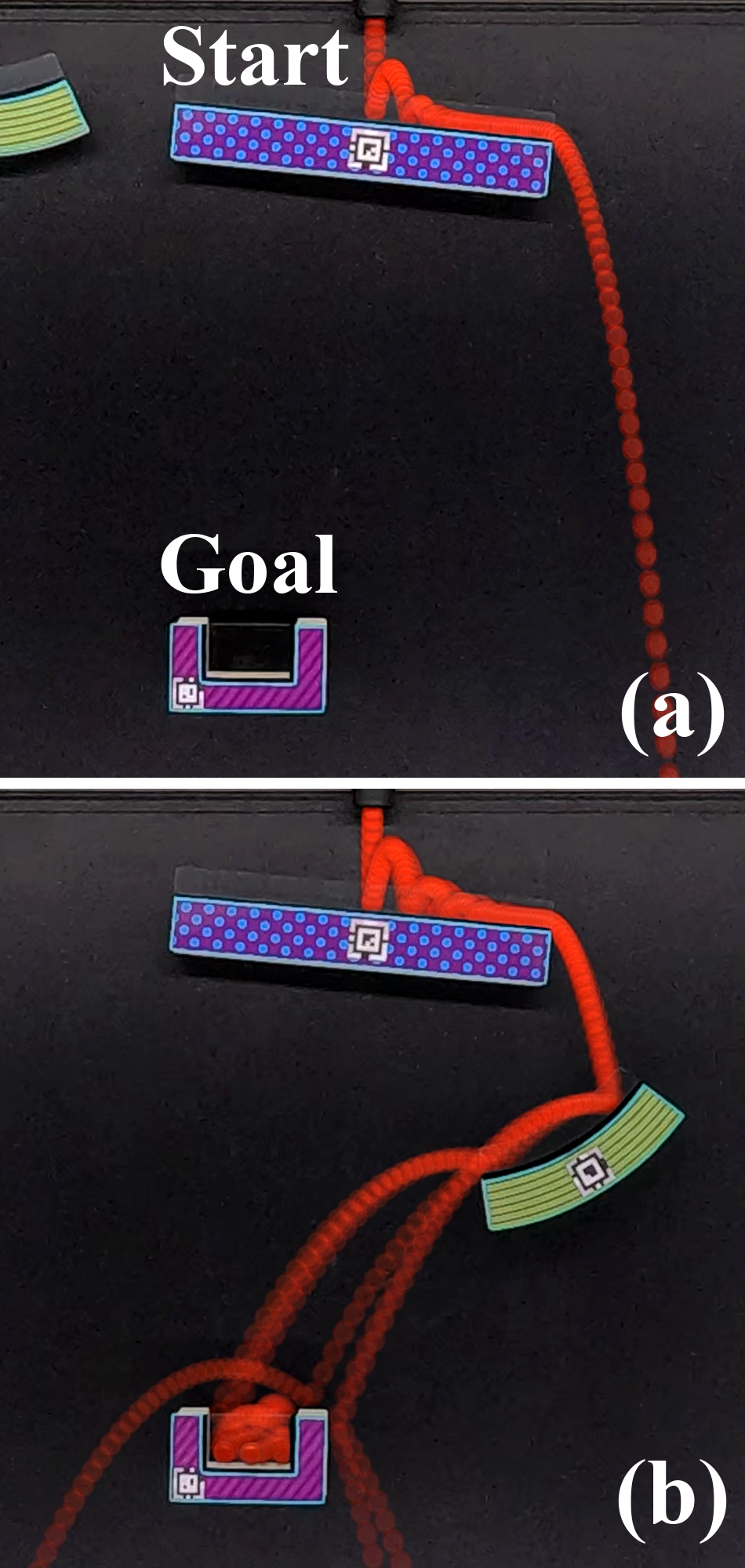}
    \end{subfigure}
\caption{The marble run robot autonomously evaluates learning algorithms on real-world marble run tasks. (a-b) An example marble run task (a) and solution (b), where the goal is to place the curved track so the ball lands in the U-shaped goal. Stochasticity of the ball's initial state leads to large variations in outcome for the same action.   }
\label{fig:robot}
\vspace{-4mm}
\end{figure}

While effective for evaluating general-purpose, long-term physical reasoning, simulation-based benchmarks neglect properties of real-world systems such as noisy observations and environment stochasticity that make reasoning difficult. One contribution of this work is the development of a robot system to enable evaluation of learning algorithms in a real-world marble run environment, shown in Fig. \ref{fig:robot}. Allowing actions only at the beginning of the task, marble run tasks are similar to simulation-based physical reasoning benchmarks but also incorporate real-world stochasticity which can cause varying outcomes for even the same initial state. Additionally, while our robot runs hundreds of trials without human intervention,  it can take up to a minute to setup and execute a single trial, so learning algorithms evaluated in this domain must perform well under a limited evaluation budget. 

Due to small changes having major effects on task outcome and the real time duration of marble run tasks, we focus on learning a structured exploration policy in simulation that can be efficiently adapted in the real world. Directly learning a deterministic policy is difficult due to this parameter sensitivity and the sim-to-real gap. We choose a policy representation that is stochastic to support exploration and captures multiple types of solutions in case the strategy that is optimal in simulation is not applicable in the real world. Specifically, we use a mixture of experts (MoE) policy representation that is composed of multiple Gaussian sub-policies and a high-level selection policy and represents multiple strategies to achieve the same or similar goals. Our proposed approach extends advantage-weighted regression \cite{peng_awr2019, nair2020awac} to train a mixture of experts policy from simulated experiences. While we do not expect the mixture policy trained offline in simulation to transfer perfectly to the real world, it should provide a good starting point to perform an online search for solutions. The sub-policies represent promising regions of the action space to explore and the high-level selection policy directs the search towards high-reward regions. Our experiments show that online learning successfully fine-tunes the mixture of experts policy within a few dozen attempts in the real world and even exceeds human performance on a test task. Our results demonstrate that representing multiple strategies promotes efficient adaptation in new environments. Strategies learned in simulation or under different dynamics can still provide useful information about where to look for solutions.

\section{RELATED WORK}

Many environments for learning physical reasoning have been explored, ranging from physics-based computer games \cite{Mnih2015HumanlevelCT,fragkiadaki2016billiards} to physical reasoning benchmarks \cite{allen2020tools,bakhtin2019phyre,Johnson_2017_CVPR,DBLP:journals/corr/abs-1910-04744}. For evaluating real-world physical reasoning capabilities beyond prediction and question answering, the most common application is contact-rich manipulation tasks \cite{agrawal2016poking,ajay2019sain,zeng2019tossingbot}. In these domains robots usually take actions and receive feedback at every time-step, which allows re-planning throughout a task and reduces the effects of uncertainty or error. The simulation-based physical reasoning environments Tools \cite{allen2020tools} and PHYRE \cite{bakhtin2019phyre} allow actions only at the start of a task and are effective benchmarks for general-purpose, long-term physical reasoning. Both PHYRE and Tools define tasks similar to marble run tasks, which all require placing an object in a scene so it interacts with other objects to reach a desired goal state. Unlike the tasks in PHYRE and Tools, our real-world marble run tasks incorporate challenges of the real world including noisy observations and environment stochasticity. 

We extend advantage-weighted regression (AWR) to mixture of expert policies. AWR formulates a constrained policy search problem as weighted supervised regression on the actions, allowing the policy to be easily updated with both online data and offline data \cite{peng_awr2019,nair2020awac}. An earlier instantiation of this framework incorporated a similar advantage-weighted policy update \cite{neumann_awr2008}. Relative entropy policy search (REPS) \cite{peters2010relative} and maximum a posteriori policy optimization (MPO) \cite{abdolmaleki2018maximum} are closely related and similarly derived as a constrained policy search, but using the dual formulation for optimizing constrained objective functions. 

Policy hierarchies in robot learning are often represented as options, which are temporally extended actions \cite{sutton1999between,bacon2017option}. In our work, actions are only applied at the start of the task and options become a set of initial actions which we represent as a mixture of experts \cite{yuksel2012moe}.
Hierarchical extensions to both REPS and MPO  have been developed \cite{daniel2016hireps,wulfmeier2019compositional} and our approach is similar reflecting the underlying similarities between REPS, MPO, and AWR. They focus on learning hierarchies incrementally from scratch which is hard and requires imposing additional constraints to learn distinct and diverse sub-policies. We formulate the problem in a simpler way by assuming access to a datatset of prior experiences to initialize the mixture policy using batched supervised regression. 
We show that a simulator generates useful training data in this domain to learn the mixture distribution's underlying structure, while online learning adapts the sub-policies and distribution over policies to a specific task or environment.

\section{THE MARBLE RUN ENVIRONMENT}

To demonstrate the challenges associated with the marble run environment, we define a task which initially consists of a rectangular track at the top of the environment and a U-shaped goal in the bottom half of the environment. In the initial configuration, a ball released at the top of the environment above the rectangular track will miss the goal, as shown in Fig. \ref{fig:robot}a. The robot must then find a configuration of the curved track that allows the ball to land in the goal. While this task appears easy given that many humans could find a solution within just a few attempts, Fig. \ref{fig:robot}b shows that, due to slight variations in the environment, an action that is successful once may not be successful every time. 

\begin{wrapfigure}{r}{0.5\linewidth}
         \includegraphics[width=\linewidth]{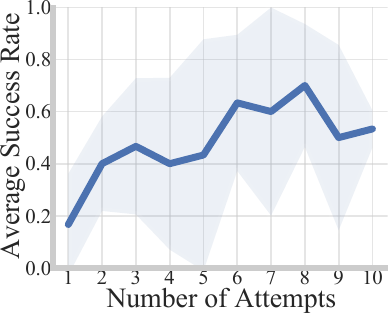}
    \caption{Average performance of 5 humans on a marble run task over 10 attempts. }
\label{fig:human_sc}
    \vspace{-1mm}

\end{wrapfigure}
Preliminary experiments demonstrate that even for a simple task, finding solutions that are robust to stochasticity in the environment is challenging even for humans. Fig. \ref{fig:human_sc} shows the average performance of 5 humans who were asked to make 10 attempts to solve a marble run task. Initially, all participants were able to find actions that were somewhat successful. Small adjustments to the initial object placement usually produced slightly improved performance, but nearly all participants eventually required switching strategies with more drastic changes to the object placement or angle in order to find more robust actions. We task the robot with finding actions that are \textit{always} or \textit{nearly always } successful, rather than just finding actions that worked once. 

\section{FRAMEWORK FOR STRUCTURED EXPLORATION}
\subsection{Problem Definition}
We focus on the problem of placing an additional track in an existing configuration so that the ball lands in the goal. A marble run task is defined by the initial configuration of all objects in the scene, including the ball, goal, and additional track pieces. In this paper, we simplify the problem by specifying the single track piece to be moved. 
The full 3D state of the environment is not observable, so we approximate the environment as 2D and ignore small out-of-plane movement. The state $\mathbf{s}$ is the initial scene configuration, which concatenates each object's $x, y$ position and orientation expressed as $[\sin(\theta),\cos(\theta)]$. Object positions are computed relative to the ball's initial position, which is fixed across tasks and therefore not included in the state. We consider tasks with the same number and types of objects so it is not necessary to include object type in the state.  
With tracks magnetically attached to a panel, only the ball is considered dynamic. 
The action $\mathbf{a}$ is the $x, y$ position and orientation of the single moved track piece. If different types of objects could be placed, the object type should also be included in the action. For now, we consider only one object type for the action.

\textbf{\textit{Reward Function}}. 
The reward function for a single trial is $ R(\mathbf{s},\mathbf{a})= \{ 1, \ \ \text{if } \ success; \ \ -d_{min} \ \ \text{otherwise}\}$, 
where $d_{min}$ is the minimum distance between the ball and the goal along the ball's trajectory. The reward is $1$ for successful trials to differentiate between true successes and trials where the ball bounces out of the goal, where $d_{min}=0$ in both cases. Uncertainty in the real system's initial state can cause the outcome to vary significantly across trials from the same state and action, so the reward is averaged over 6 trials per action taken in the real world or in a stochastic simulator. We empirically found 6 trials provides a good trade-off between minimizing evaluation time and minimizing variance. 
\subsection{Mixture of Experts (MoE) Policy Representation}
To represent knowledge learned from past experiences, we learn a stochastic policy parameterized as a probabilistic mixture of experts (MoE) that maps an input state to a multi-modal distribution of actions. The MoE consists of $K$ Gaussian ``expert'' policies $\{\pi_k\}_{ \forall k \in \{1,\dots, K \} }$ and a ``gating'' policy $\psi$ that predicts a categorical distribution over the experts such that 
\begin{align}
    k &\sim \text{Categorical}(\psi(k|\mathbf{s},\boldsymbol{\phi})) \label{psi}\\
    \mathbf{a} &\sim \pi_k(\mathbf{a}|\mathbf{s},\boldsymbol{\theta}_k), \label{pi_k}
\end{align}
where $\boldsymbol{\phi}$ and $\boldsymbol{\theta}_k$ are learned parameters for the neural networks representing $\psi$ and $\pi_k$. $k$ is the expert model index sampled from the categorical distribution predicted by $\psi$ and $\mathbf{a}$ is the action sampled from the Gaussian distribution with mean $\boldsymbol{\mu}_k$ and covariance $\boldsymbol{\Sigma}_k$ represented by $\theta_k$.

\subsection{Learning a Mixture of Experts Policy from (Simulated) Experience}
\label{sec:pretrain}
We generate a dataset of prior experiences using a simulated marble run environment by randomly sampling actions on a set of training tasks until 500 successful and unsuccessful actions are found for each task. Each sample is stored in the dataset $\mathcal{D}$ as a state-action-reward tuple $(\mathbf{s},\mathbf{a},R(\mathbf{s},\mathbf{a}))$. We do not expect the mixture policy trained in simulation to transfer perfectly to the real world, but it should provide a good starting point to search for potential solutions. 


\textbf{\textit{Expectation Maximization for Mixture of Experts Policies.}}
We derive a training procedure from the expectation maximization (EM) algorithm \cite{ng2004em}. The EM algorithm estimates the parameters  $\boldsymbol{\phi}$ and $\boldsymbol{\theta} = \{\boldsymbol{\theta}_k\}_{k\in1:K}$ that maximize the complete log-likelihood of the selection variables $\{z_k\}$ and actions $\mathbf{a}$ given the state $\mathbf{s}$. The selection variable $z_k$ is $1$ if the $k^{th}$ expert generates or predicts the action  $\mathbf{a}$ and $0$ otherwise. The E-step calculates the expected log-likelihood, given as $J(\boldsymbol{\phi},\boldsymbol{\theta})$ below, where $w'_k$ is the probability $z_k$ is one given the state and action.
\begin{align}
\begin{split}
\label{eq:em_obj}
J(\boldsymbol{\phi},\boldsymbol{\theta}) = \underset{\mathbf{s},\mathbf{a} \sim \mathcal{D}}{\mathbb{E}}\Bigg[\sum^K_{k=1}  w'_k \Big(\log & \pi_k(\mathbf{a}|\mathbf{s},\boldsymbol{\theta}_k) \\
&+ \log \psi_k(\mathbf{s},\boldsymbol{\phi})\Big)\Bigg]\\ 
\end{split}
\end{align}
\begin{align}
\label{eq:resp}
    w'_k = P(z_k=1|\mathbf{s},\mathbf{a}) = \frac{\psi_k(\mathbf{s},\boldsymbol{\phi}')\pi_k(\mathbf{a}|\mathbf{s},\boldsymbol{\theta}'_k)}{\sum^K_{j=1}\psi_j(\mathbf{s},\boldsymbol{\phi}')\pi_j(\mathbf{a}|\mathbf{s},\boldsymbol{\theta}'_j)}
\end{align}
The superscript $\ ' \ $ indicates $w_k$ is computed using the current parameter estimates and is not involved in the gradient calculation.
In the standard EM algorithm, the M-step analytically computes parameters which maximize $J(\boldsymbol{\phi},\boldsymbol{\theta})$. There is not a closed-form solution when $\psi$ and $\pi_k$ are neural networks with non-linear activations as in this work so we instead use a generalized EM algorithm, where the M-step performs a gradient step to move $J(\boldsymbol{\phi},\boldsymbol{\theta})$ closer to the maximum \cite{Neal1998AVO}. 

\textbf{\textit{Advantage-Weighted Regression.}}
Using the mixture log-likelihood in  (\ref{eq:em_obj}), we apply advantage-weighted regression (AWR), which weights the log-likelihood with the exponential advantage $\exp(\frac{ A^{\pi}(\mathbf{s},\mathbf{a})}{\eta})$ \cite{peng_awr2019,nair2020awac,neumann_awr2008}.
The advantage $A^{\pi}(\mathbf{s},\mathbf{a}) = R(\mathbf{s},\mathbf{a}) - V^{\pi}(\mathbf{s})$ is a measure of improvement based on how the reward of an action compares to the average reward observed from a state under the current policy. When the advantage is negative, the weights approach zero and  filter out poorly performing actions. The resulting advantage-weighted objective function is 
\begin{align}
\begin{split}
J(\boldsymbol{\phi},\boldsymbol{\theta}) =  \underset{\mathbf{s},\mathbf{a} \sim \mathcal{D}}{\mathbb{E}}\Bigg[&\sum^K_{k=1} w'_k \exp\left(\frac{A^{\pi'_k}(\mathbf{s},\mathbf{a})}{\eta} \right)\\
&\Big(\log \pi_k(\mathbf{a}|\mathbf{s},\boldsymbol{\theta}_k) + \log \psi_k(\mathbf{s},\boldsymbol{\phi})\Big) \Bigg]\label{eq:em_obj_adv},
\end{split}
\end{align}
where $\eta$ is a Lagrange multiplier associated with constraining the policy to stay close to the behavior policy $\pi_{\beta}$ that represents the distribution of data seen so far. Dual gradient descent can be used to estimate $\eta$, but this requires estimating $\pi_{\beta}$ from the data \cite{wulfmeier2019compositional}. 
We treat $\eta$ as a fixed hyperparameter, which has been effective in prior work  \cite{peng_awr2019,nair2020awac}. 
In marble run tasks, the final episode reward is observed immediately after taking each action, so the value function is the average reward of actions sampled from the current policy at a specified state. During offline training, we pre-compute the per-state values as the average reward of actions observed in the dataset at each state.

\subsection{Online Learning}
\label{sec:finetune}
The mixture policy trained offline in simulation represents prior knowledge about what strategies might work well in different contexts. When new environments or task configurations are encountered, the offline policy is a starting point from which to perform an online search for robust solutions in the current context. Expert policies represent promising regions of the action space to explore. The gating network directs the search towards the most promising regions.   

\textbf{\textit{Decomposing the Objective Function with Hard Policy Updates.}} As a supervised learning algorithm, advantage-weighted regression is easily adapted to online learning by incorporating online samples into policy updates. The objective function defined in  (\ref{eq:em_obj_adv}) requires indiscriminately optimizing all sub-policies over batches sampled across the entire dataset with each update step, where samples are weighted according to responsibilities  $w'_k$. This soft policy update shares information between policies and is important during early stages of training to learn a better division of the state space. We empirically observe that after pre-training the above soft updates become less effective during online learning. We assume this is because a locally optimal division of the state space is learned during pre-training.

To compensate for the declining effectiveness of soft updates, we perform hard policy updates during online learning by updating expert policies independently using only data associated with each policy.  Performing policy updates using only the most relevant samples helps sub-policies quickly specialize to the current task. 
The individual update rules can be derived by setting $w'_k$ to $0$ or $1$ in  (\ref{eq:em_obj_adv}), where $1$ is assigned to the component with the highest probability $P(z_k=1|\mathbf{s},\mathbf{a})$. We decompose  (\ref{eq:em_obj_adv}) into separate objective functions for the gating policy and each expert policy. The objective function for policy $k$ is 
\begin{align}
J(\boldsymbol{\theta}_k) =  \underset{\mathbf{s},\mathbf{a} \sim \mathcal{D}_k}{\mathbb{E}}\Big[\exp\left(\frac{A^{\pi'_k}(\mathbf{s},\mathbf{a})}{\eta} \right)\log \pi_k(\mathbf{a}|\mathbf{s},\boldsymbol{\theta}_k) 
\Big]\label{eq:em_obj_adv_expert},
\end{align}
where $\mathcal{D}_k$ is a subset of the dataset generated by or associated with the $k^{th}$ policy. We also decompose the advantage function. The advantage function for the $k^{th}$ policy is defined as $A^{\pi'_k}(\mathbf{s},\mathbf{a}) = R_{\boldsymbol{\omega}_k}(\mathbf{s},\mathbf{a}) - \mathbb{E}_{\mathbf{a}\sim\pi'_k}[R_{\boldsymbol{\omega}_k}(\mathbf{s},\mathbf{a})]$. Similarly, the gating policy's objective function is
\begin{align}
J(\boldsymbol{\psi}) =  \underset{\mathbf{s},k \sim \mathcal{D}}{\mathbb{E}}\Big[ \exp\left(\frac{A^{\psi'}(\mathbf{s},k)}{\eta} \right)\log \psi_k(\mathbf{s},\boldsymbol{\phi}) 
\Big]\label{eq:em_obj_adv_gate},
\end{align}
and its advantage function is $ A^{\psi'}(\mathbf{s},k) = \mathbb{E}_{\mathbf{a}\sim\pi'_k}[R_{\boldsymbol{\omega}_k}(\mathbf{s},\mathbf{a})] - \mathbb{E}_{k\sim\psi'}[\mathbb{E}_{\mathbf{a}\sim\pi'_k}[R_{\boldsymbol{\omega}_k}(\mathbf{s},\mathbf{a})]]$.
The gating policy's advantage function provides an estimate of how actions from expert $k$ compare to other experts. A similar advantage function is defined in hierarchical reinforcement learning as the advantage over options for determining option termination criteria \cite{bacon2017option}.

\textbf{\textit{Learning an Approximate Reward Function.}}
 During online learning, the advantage is estimated using learned reward functions. Learning a single function to approximate a multi-modal, discontinuous reward function is difficult, so different learned reward functions are associated with each expert policy. Each approximate reward function $R_{\boldsymbol{\omega}_k}(\mathbf{s},\mathbf{a})$ is trained using only data generated by the associated expert. 
During offline training of the policy, the best divisions of the state-action space are not yet known, so we estimate the advantage directly from sampled rewards to avoid bias. After the mixture policy is trained offline, we perform a hard assignment of samples to the most likely policy indicated by responsibilities $w'_k$ and  $R_{\boldsymbol{\omega }_k}(\mathbf{s},\mathbf{a})$ is pre-trained over the corresponding subset of data, $\mathcal{D}_k$, by minimizing the mean-squared-error (MSE) between predicted and observed rewards. An equal number of positive and negative samples is used in each update for both pre-training the reward functions and during online learning. Samples are considered positive if at least half the trials for the action are successful (i.e. success rate $\geq$ 0.5).
During online learning, each reward function is updated using both online and offline samples from the corresponding policy.

\textbf{\textit{Summary of Online Learning Algorithm.}} We assume the online learning phase for each new task is initialized with the same offline policy and dataset. At each iteration of online learning, an expert policy and action are sampled from the mixture policy ($k\sim\psi,\ \mathbf{a} \sim \pi_k$). The action is executed and stored in the dataset as a tuple $(\mathbf{s},\mathbf{a},k,R(\mathbf{s},\mathbf{a}))$. The learned reward function $R_{\boldsymbol{\omega}_k}(\mathbf{s},\mathbf{a})$ associated with the current expert $k$ is then updated with a batch of training points $(\mathbf{s}_k,\mathbf{a}_k,R(\mathbf{s}_k,\mathbf{a}_k)) \sim \mathcal{D}_k$ containing an even mixture of positive and negative samples. The policy $\pi_k$ is updated with a batch of training points $(\mathbf{s}_k,\mathbf{a}_k,R(\mathbf{s}_k,\mathbf{a}_k)) \sim \mathcal{D}_k$ using  (\ref{eq:em_obj_adv_expert}) and the gating policy $\psi$ is updated using  (\ref{eq:em_obj_adv_gate}) with a batch of training points $(\mathbf{s},\mathbf{a},k,R(\mathbf{s},\mathbf{a})) \sim \mathcal{D}$ sampled over the entire dataset. The batches used for the policy and reward function updates are composed of a balanced ratio of online-to-offline samples, providing more aggressive updates than uniform sampling. The ratio is initially set to 0 and linearly increased to 1 over $N$ steps, where each batch contains only online samples by the $N^{th}$ step. For the expert policy and reward function updates, we increase the ratio to 1 over 25 steps. For the gating network, we increase the ratio more slowly over 100 steps to preserve exploration and reduce the risk of premature convergence to a sub-optimal strategy.  

\textbf{\textit{Additional Implementation Details.}} The sub-policies, gating network, and reward function are all represented as multi-layer perceptions (MLP) with 2 hidden layers, where each layer has 256 units and is followed by ReLU activations. For the sub-policies, the output layer splits into two heads for the mean and covariance. To estimate the covariance, the network outputs Cholesky factors  $A$ such that $\Sigma = AA^T$, where $A$ is lower triangular and the diagonals of $A$ are $A_{ii} \leftarrow \exp(A_{ii}) + \epsilon$. The Gumbel-Softmax activation, which provides a  differentiable approximation of samples drawn from a categorical distribution \cite{jang2016categorical}, is applied to the output layer of the gating network. All networks are optimized using the Adam optimizer.
\nopagebreak[4]

\section{Evaluation}

\begin{figure*}[t!]
\begin{subfigure}[t]{0.34\linewidth}
     \includegraphics{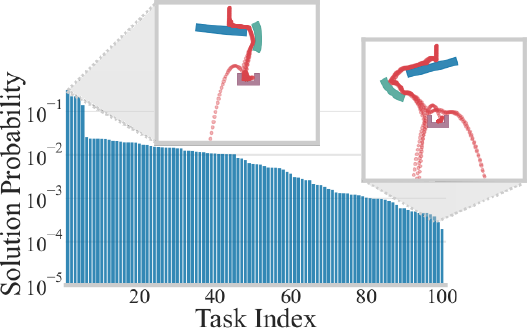}
  \caption{}
        \label{fig:sol_probs}
\end{subfigure}
\begin{subfigure}[t]{0.65\linewidth}
     \includegraphics{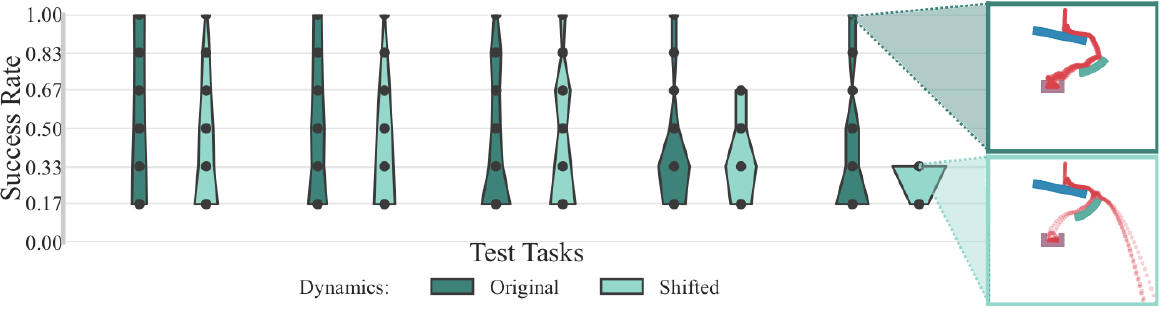}
     \caption{}
     \label{fig:task_success_distr}
     
\end{subfigure}
  \caption{(a) Solution probabilities per task estimated over 10,000 random actions in simulation. Tasks increase in difficulty from left to right. (b) Success rate distribution of actions with success rate $>$ 0 out of 10,000 randomly sampled actions. Results are shown for 5 test tasks, sorted in order of increasing difficulty. The width at each point corresponds to the proportion of occurrences with a success rate of that value. Results are also shown for different simulation dynamics (shifted) where we apply a horizontal wind-like force in the same direction the ball rolls off the initial rectangular track. Depending on the environment dynamics, it may be more difficult to find actions with high success rates for some tasks.
  }
  \vspace{-4mm}
 \end{figure*}
 
\subsection{Simulation Setup}
The simulated marble run environment is built in Box2D (\href{https://box2d.org/}{box2d.org}) using 2D models of the real marble run tracks extracted from RGB images. The physical parameters (coefficient of restitution, friction, and gravity are optimized to match data from the real system. The ball's initial position and velocity are assumed fixed but do vary in the real world because the ball's diameter does not match the diameter of the launching tube and its velocity is not explicitly controlled. Noise is added to the ball's initial state to reflect this stochasticity. For some experiments, we add an additional horizontal gravitation force which acts like wind to represent a shift in dynamics from the training environment to the test environment. We add this force in the same direction that the ball rolls off the rectangular track to prevent cases where wind slows the ball to a stop and the task become unsolvable. 

\subsection{Metrics}
  The average success rate is used as a performance metric, where the success rate refers to the number of successful trials out of 6 taken for each action.  We find aggregating performance across tasks computed using the arithmetic mean can be dominated by outlier tasks (i.e. very easy or very difficult tasks). 
  We use the inter-quartile mean (IQM),  which is less sensitive to outliers and stratified bootstrap confidence intervals to report aggregate performance~\cite{agarwal2021deep}.

\subsection{Task Dataset Generation}
We randomly generate 100 marble run tasks with varying initial configurations of a long rectangular track and a U-shaped goal. The objective is to place a curved track so the ball lands in the goal. On the real system, the ball is dropped through a fixed tube so we assume its mean initial position and velocity are the same for every task.  The rectangular track is placed near the tube to catch the ball, but varies slightly in $x$, $y$, and $\theta$. The goal position is more varied, where the range of $x$ nearly spans the environment width, the range of $y$ spans the region below the rectangular track, and  its angle is always $0$. We use task generation scripts from the PHYRE code-base to ensure tasks are non-trivial and sufficiently diverse \cite{bakhtin2019phyre}. The tasks are split into 80 training, 10 validation, and 10 test tasks.

\textbf{\textit{Difficulty of Marble Run Tasks}}. We evaluate task difficulty using the stochastic simulation environment by estimating the solution probability for each task as the average success rate of 10,000 randomly sampled actions. The solution probabilities shown in Fig. \ref{fig:sol_probs}  demonstrate varying difficulty across tasks, with some requiring more than 10,000 actions before finding a solution with random sampling alone.

The average success rate that can be achieved for each task depends on the environment  dynamics and may be considerably less than 1 on difficult tasks, which is shown in Fig. \ref{fig:task_success_distr}. When a wind-like force is introduced in the environment, lower success rates under shifted dynamics indicate some tasks are more difficult to solve. If finding actions with success rates of 1 is possible, such actions may be rare, difficult to reproduce, or occur by chance. Estimating the success rate with more than 6 trials per action would reduce the occurrence of finding high success rate actions by chance, but with increased run-time. 
  \begin{figure*}[t!]
  \centering
      \begin{subfigure}{\linewidth}
              \includegraphics[width=\linewidth]{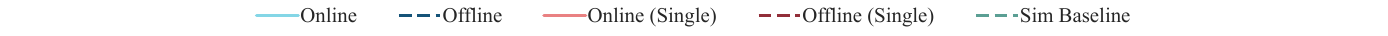}
     \end{subfigure}\\
    \begin{subfigure}[t]{0.49\linewidth}
         \includegraphics[width=0.5\textwidth]{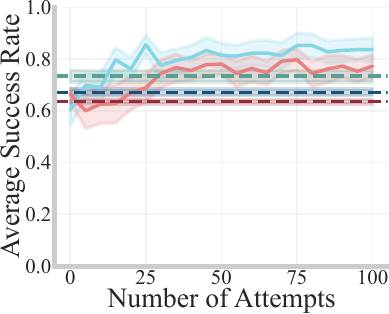}
         \includegraphics[width=0.49\textwidth]{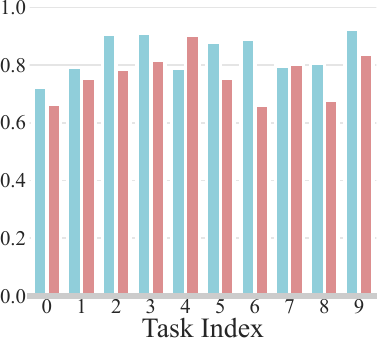}
         \caption{Simulation}
                           \label{fig:sim_results_sim}

     \end{subfigure}
     \begin{subfigure}[t]{0.49\linewidth}
         \includegraphics[width=0.5\textwidth]{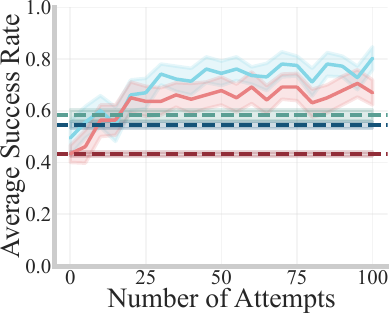}
         \includegraphics[width=0.49\textwidth]{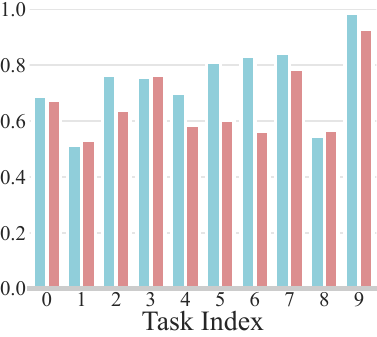}
        \caption{Simulation with Wind}
        \label{fig:sim_results_shifted}
     \end{subfigure}
      \caption{ Online learning performance in  (a) simulation and (b) simulation with a wind-like force producing different dynamics than the offline dataset. Average success rate vs. number of attempts (per task) is shown for the evaluation steps taken every 5 attempts during online learning. The bar charts show the average success rate over all evaluation steps for each task.
}
      \label{fig:sim_results}
      \vspace{-2mm}
  \end{figure*}
\subsection{Method Comparisons}
\textbf{Offline Mixture of Experts [Offline]:} We evaluate the mixture of experts policy's performance after training offline on the simulated dataset, as described in Section \ref{sec:pretrain}. Offline performance is reported as the average success rate of 20 actions sampled from the mixture policy and evaluated on the test tasks. Actions are selected by sampling an expert policy from the categorical distribution over policies (i.e. the gating network) and then using the mean of the sampled policy as the action to evaluate. 
\textbf{Online Mixture of Experts [Online]:}  The mixture of experts policy is updated with online learning, as described in Section \ref{sec:finetune}, by attempting 100 actions and updating the policy after each attempt. Every 5 attempts, we take an additional action to evaluate the mean of the current expert policy and record the performance. These evaluation actions are used only to report performance and are not used in the policy update. 
\textbf{Simulation Performance Baseline [Sim Baseline]:} As a performance baseline for simulation-based experiments, we rank 10,000 randomly sampled actions for each test task using a perfect model of the evaluation environment (i.e. the simulator). The average success rate of the top 5 actions ranked by the model is reported. As the number of sampled actions approaches infinity, the baseline performance would represent the best performance that could possibly be achieved on the simulated test tasks.  The baseline's reported performance may be lower than the true best performance because it is limited to ranking 10,000 actions per task.   \textbf{Single Gaussian Policy [Single]:} To emphasize the importance of representing multiple strategies, we compare the MoE policy performance to that of a single Gaussian policy trained using the same procedures as in Sections \ref{sec:pretrain} and \ref{sec:finetune}, except only a single policy is used so no gating network is learned.

\subsection{Simulation Experiments}

In Fig. \ref{fig:sim_results}, we show the average success rate of evaluation steps taken every 5 attempts during online learning in simulation (Fig. \ref{fig:sim_results_sim}) and with different simulated dynamics than the offline dataset (Fig. \ref{fig:sim_results_shifted}). The bar charts show the per-task success rates averaged over all evaluation steps. 

When the dynamics of the training and test environments match, the performance gained by representing multiple strategies is less pronounced. The offline performance of the single Gaussian policy falls slightly below the MoE policy, likely due to the single Gaussian policy averaging over multiple solution regions which may include pockets of lower reward regions between them. The MoE policy can represent distinct solution regions as separate policies which allows the policies to fit more closely to the high reward regions and converge more quickly during online learning.

The benefits of representing multiple strategies are more easily observed when the dynamics of the test environment do not match the dynamics of the training environment, as in Fig. \ref{fig:sim_results_shifted}. 
At the start of convergence, after around 25 attempts, the MoE continues to increase beyond the performance of the single Gaussian policy. This indicates that the MoE policy is more capable of escaping local optima by switching between different candidate solutions. Escaping local optima is especially important under different dynamics, because the best strategies for a task will change depending on the dynamics. The relative robustness of the MoE policy to shifts in dynamics is further shown by the offline policy performance which is less affected by the shifted dynamics than the single Gaussian policy. 
Please visit \href{https://sites.google.com/view/learning-strategies-icra2023/home}{https://sites.google.com/view/learning-strategies-icra2023/home} to view supplementary videos.
\subsection{Real-World Experiments}

  \begin{figure}
  \centering
   \begin{subfigure}{\linewidth}
              \includegraphics[width=\linewidth]{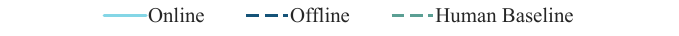}
     \end{subfigure}\\
     \begin{subfigure}{0.49\linewidth}
\includegraphics{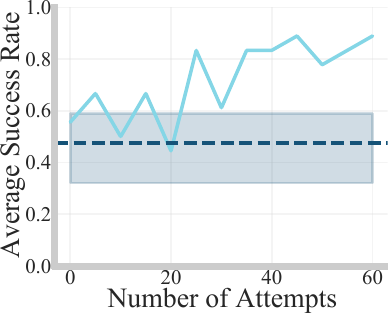}
\caption{Real World}
     \end{subfigure}
          \begin{subfigure}{0.49\linewidth}
\includegraphics{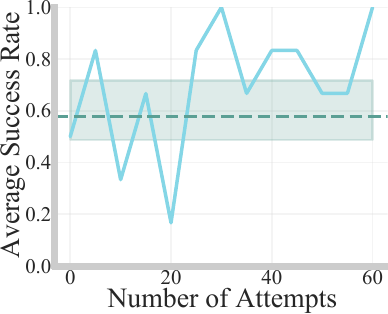}
\caption{Test Task 7 }
     \end{subfigure}
\caption{(a) Online learning performance in the real world evaluated on a subset of 5 test tasks. (b) Comparison to human performance for a single test task. }
\label{fig:real_results}
\vspace{-5mm}
  \end{figure}
Fig. \ref{fig:real_results}a shows offline and online MoE policy performance on the real system, evaluated on a random subset of 5 test tasks. We also limit online learning to just 60 attempts to reduce run-time. Despite the mismatched dynamics between the simulation environment and the real world, the MoE policy achieves an average success rate just over 0.8 within a few dozen attempts, which is consistent with the simulation results.  In Fig. \ref{fig:real_results}b, we compare the performance of the MoE policy to the human performance from Fig. \ref{fig:human_sc} which was evaluated on test task 7.  We plot the average performance from the last 5 attempts as the human baseline.  The MoE policy starts off with around the same performance as the human baseline, but eventually exceeds human performance. By the end of online learning, the average performance of the MoE policy is hovering between 0.7 and 1 so the asymptotic performance is likely in the range of 0.8-0.9 for that task.   
\section{CONCLUSIONS}



We present a method using a mixture of experts policy to represent multiple strategies for solving marble run tasks.
Our experiments demonstrate that, even when trained offline on simulated data, online learning quickly adapts the policy to solve new marble run tasks in the real world. 
Finally, by developing a robot system to evaluate the proposed approach on real-world marble run tasks, this work emphasizes the importance of enabling experimental evaluation in domains that involve complex dynamic interactions in the physical world. 








\bibliographystyle{IEEEtran}
\bibliography{IEEEabrv,references}

\begin{thebibliography}{10}
\providecommand{\url}[1]{#1}
\csname url@rmstyle\endcsname
\providecommand{\newblock}{\relax}
\providecommand{\bibinfo}[2]{#2}
\providecommand\BIBentrySTDinterwordspacing{\spaceskip=0pt\relax}
\providecommand\BIBentryALTinterwordstretchfactor{4}
\providecommand\BIBentryALTinterwordspacing{\spaceskip=\fontdimen2\font plus
\BIBentryALTinterwordstretchfactor\fontdimen3\font minus
  \fontdimen4\font\relax}
\providecommand\BIBforeignlanguage[2]{{%
\expandafter\ifx\csname l@#1\endcsname\relax
\typeout{** WARNING: IEEEtran.bst: No hyphenation pattern has been}%
\typeout{** loaded for the language `#1'. Using the pattern for}%
\typeout{** the default language instead.}%
\else
\language=\csname l@#1\endcsname
\fi
#2}}

\bibitem{allen2020tools}
\BIBentryALTinterwordspacing
K.~R. Allen, K.~A. Smith, and J.~B. Tenenbaum, ``Rapid trial-and-error learning
  with simulation supports flexible tool use and physical reasoning,''
  \emph{Proceedings of the National Academy of Sciences}, vol. 117, no.~47, pp.
  29\,302--29\,310, 2020. [Online]. Available:
  \url{https://www.pnas.org/content/117/47/29302}
\BIBentrySTDinterwordspacing

\bibitem{bakhtin2019phyre}
A.~Bakhtin, L.~van~der Maaten, J.~Johnson, L.~Gustafson, and R.~Girshick,
  ``Phyre: A new benchmark for physical reasoning,'' in \emph{Advances in
  Neural Information Processing Systems}, vol.~32, 2019.

\bibitem{peng_awr2019}
X.~B. Peng, A.~Kumar, G.~Zhang, and S.~Levine, ``Advantage-weighted regression:
  Simple and scalable off-policy reinforcement learning,'' \emph{arXiv preprint
  arXiv:1910.00177}, 2019.

\bibitem{nair2020awac}
\BIBentryALTinterwordspacing
A.~Nair, M.~Dalal, A.~Gupta, and S.~Levine, ``Accelerating online reinforcement
  learning with offline datasets,'' \emph{CoRR}, vol. abs/2006.09359, 2020.
  [Online]. Available: \url{https://arxiv.org/abs/2006.09359}
\BIBentrySTDinterwordspacing

\bibitem{Mnih2015HumanlevelCT}
V.~Mnih, K.~Kavukcuoglu, D.~Silver, A.~A. Rusu, J.~Veness, M.~G. Bellemare,
  A.~Graves, M.~A. Riedmiller, A.~Fidjeland, G.~Ostrovski, S.~Petersen,
  C.~Beattie, A.~Sadik, I.~Antonoglou, H.~King, D.~Kumaran, D.~Wierstra,
  S.~Legg, and D.~Hassabis, ``Human-level control through deep reinforcement
  learning,'' \emph{Nature}, vol. 518, pp. 529--533, 2015.

\bibitem{fragkiadaki2016billiards}
K.~Fragkiadaki, P.~Agrawal, S.~Levine, and J.~Malik, ``Learning visual
  predictive models of physics for playing billiards,'' in \emph{ICLR}, 2016.

\bibitem{Johnson_2017_CVPR}
J.~Johnson, B.~Hariharan, L.~van~der Maaten, L.~Fei-Fei, C.~Lawrence~Zitnick,
  and R.~Girshick, ``Clevr: A diagnostic dataset for compositional language and
  elementary visual reasoning,'' in \emph{Proceedings of the IEEE Conference on
  Computer Vision and Pattern Recognition (CVPR)}, July 2017.

\bibitem{DBLP:journals/corr/abs-1910-04744}
\BIBentryALTinterwordspacing
R.~Girdhar and D.~Ramanan, ``{CATER:} {A} diagnostic dataset for compositional
  actions and temporal reasoning,'' \emph{CoRR}, vol. abs/1910.04744, 2019.
  [Online]. Available: \url{http://arxiv.org/abs/1910.04744}
\BIBentrySTDinterwordspacing

\bibitem{agrawal2016poking}
\BIBentryALTinterwordspacing
P.~Agrawal, A.~Nair, P.~Abbeel, J.~Malik, and S.~Levine, ``Learning to poke by
  poking: Experiential learning of intuitive physics,'' \emph{CoRR}, vol.
  abs/1606.07419, 2016. [Online]. Available:
  \url{http://arxiv.org/abs/1606.07419}
\BIBentrySTDinterwordspacing

\bibitem{ajay2019sain}
A.~Ajay, M.~Bauza, J.~Wu, N.~Fazeli, J.~B. Tenenbaum, A.~Rodriguez, and L.~P.
  Kaelbling, ``{Combining Physical Simulators and Object-Based Networks for
  Control},'' in \emph{IEEE International Conference on Robotics and Automation
  (ICRA)}, 2019.

\bibitem{zeng2019tossingbot}
A.~Zeng, S.~Song, J.~Lee, A.~Rodriguez, and T.~Funkhouser, ``Tossingbot:
  Learning to throw arbitrary objects with residual physics,'' in
  \emph{Proceedings of Robotics: Science and Systems (RSS)}, 2019.

\bibitem{neumann_awr2008}
G.~Neumann and J.~Peters, ``Fitted q-iteration by advantage weighted
  regression,'' in \emph{Advances in Neural Information Processing Systems},
  D.~Koller, D.~Schuurmans, Y.~Bengio, and L.~Bottou, Eds., vol.~21.\hskip 1em
  plus 0.5em minus 0.4em\relax Curran Associates, Inc., 2009.

\bibitem{peters2010relative}
J.~Peters, K.~Mulling, and Y.~Altun, ``Relative entropy policy search,'' in
  \emph{Twenty-Fourth AAAI Conference on Artificial Intelligence}, 2010.

\bibitem{abdolmaleki2018maximum}
\BIBentryALTinterwordspacing
A.~Abdolmaleki, J.~T. Springenberg, Y.~Tassa, R.~Munos, N.~Heess, and
  M.~Riedmiller, ``Maximum a posteriori policy optimisation,'' in
  \emph{International Conference on Learning Representations}, 2018. [Online].
  Available: \url{https://openreview.net/forum?id=S1ANxQW0b}
\BIBentrySTDinterwordspacing

\bibitem{sutton1999between}
R.~S. Sutton, D.~Precup, and S.~Singh, ``Between mdps and semi-mdps: A
  framework for temporal abstraction in reinforcement learning,''
  \emph{Artificial intelligence}, vol. 112, no. 1-2, pp. 181--211, 1999.

\bibitem{bacon2017option}
P.-L. Bacon, J.~Harb, and D.~Precup, ``The option-critic architecture,'' in
  \emph{Proceedings of the AAAI Conference on Artificial Intelligence},
  vol.~31, no.~1, 2017.

\bibitem{yuksel2012moe}
S.~E. Yuksel, J.~N. Wilson, and P.~D. Gader, ``Twenty years of mixture of
  experts,'' \emph{IEEE Transactions on Neural Networks and Learning Systems},
  vol.~23, no.~8, pp. 1177--1193, 2012.

\bibitem{daniel2016hireps}
\BIBentryALTinterwordspacing
C.~Daniel, G.~Neumann, O.~Kroemer, and J.~Peters, ``Hierarchical relative
  entropy policy search,'' \emph{Journal of Machine Learning Research},
  vol.~17, no.~93, pp. 1--50, 2016. [Online]. Available:
  \url{http://jmlr.org/papers/v17/15-188.html}
\BIBentrySTDinterwordspacing

\bibitem{wulfmeier2019compositional}
\BIBentryALTinterwordspacing
M.~Wulfmeier, A.~Abdolmaleki, R.~Hafner, J.~T. Springenberg, M.~Neunert,
  T.~Hertweck, T.~Lampe, N.~Y. Siegel, N.~Heess, and M.~A. Riedmiller,
  ``Compositional transfer in hierarchical reinforcement learning,'' in
  \emph{Robotics Science and Systems}, 2020. [Online]. Available:
  \url{https://roboticsconference.org/2020/program/papers/54.html}
\BIBentrySTDinterwordspacing

\bibitem{ng2004em}
S.-K. Ng and G.~McLachlan, ``Using the em algorithm to train neural networks:
  misconceptions and a new algorithm for multiclass classification,''
  \emph{IEEE Transactions on Neural Networks}, vol.~15, no.~3, pp. 738--749,
  2004.

\bibitem{Neal1998AVO}
R.~M. Neal and G.~E. Hinton, ``A view of the em algorithm that justifies
  incremental, sparse, and other variants,'' in \emph{Learning in Graphical
  Models}, 1998.

\bibitem{jang2016categorical}
\BIBentryALTinterwordspacing
E.~Jang, S.~Gu, and B.~Poole, ``Categorical reparameterization with
  gumbel-softmax,'' in \emph{International Conference on Learning
  Representations (ICLR)}, 2016. [Online]. Available:
  \url{https://arxiv.org/abs/1611.01144}
\BIBentrySTDinterwordspacing

\bibitem{agarwal2021deep}
R.~Agarwal, M.~Schwarzer, P.~S. Castro, A.~Courville, and M.~G. Bellemare,
  ``Deep reinforcement learning at the edge of the statistical precipice,''
  \emph{Advances in Neural Information Processing Systems}, 2021.

\end{thebibliography}

\end{document}